\pdfoutput=1

\documentclass[11pt]{article}

\usepackage{EACL2023}

\usepackage{times}
\usepackage{latexsym}

\usepackage[T1]{fontenc}

\usepackage[utf8]{inputenc}

\usepackage{microtype}

\usepackage{inconsolata}

\title{Exploring    Paracrawl for Document-level Neural Machine Translation}

\author{Yusser Al Ghussin$^{1,2}$, Jingyi Zhang$^3$ and Josef van Genabith$^{1,2}$ \\
  $^1$German Research Center for Artificial Intelligence (DFKI),\\
Saarland Informatics Campus, Saarbrucken, Germany  \\
$^2$Department of Language Science and Technology, Saarland University, Germany\\
  $^3$Hasso-Plattner-Institut (HPI),  Potsdam, Germany\\
  \texttt{yusser.al\_ghussin/Josef.Van\_Genabith@dfki.de,Jingyi.Zhang@hpi.de} \\}

\begin{document}
\maketitle
\begin{abstract}

 Document-level neural machine translation (NMT) has outperformed sentence-level NMT on a number of datasets. However, document-level NMT is still not widely adopted in real-world translation systems mainly due to the lack of large-scale general-domain training data for document-level NMT. We examine the effectiveness of using Paracrawl  for 
learning document-level translation. Paracrawl is a large-scale parallel corpus crawled from the Internet and contains data from various domains. 
The
 official  Paracrawl corpus was released as    parallel sentences (extracted from parallel webpages) and therefore previous works   only used Paracrawl   for learning sentence-level translation.
 In this work, we  extract parallel paragraphs from Paracrawl parallel webpages  using automatic sentence alignments and  we use the extracted parallel paragraphs as parallel documents for training document-level translation models.
  We show that
document-level NMT models trained with only parallel paragraphs from Paracrawl  can be used to translate real documents   from  TED, News and Europarl, outperforming sentence-level NMT models. We also perform a targeted pronoun evaluation  and show that document-level models trained with Paracrawl data can help  context-aware pronoun
translation.   
We release our
data and code here\footnote{{https://github.com/Yusser96/Exploring-Paracrawl-for-Document-level-Neural-Machine-Translation}}.
\end{abstract}

\section{Introduction}

The Transformer translation model \cite{vaswani2017attention}, which performs sentence-level translation based on attention networks, has achieved great success and significantly improved the state-of-the-art in machine translation. Compared to sentence-level translation, document-level translation \cite{xu-etal-2021-document-graph,bao-etal-2021-g,jauregi-unanue-etal-2020-leveraging,ma-etal-2020-simple,maruf-etal-2019-selective,tu-etal-2018-learning,maruf-haffari-2018-document}
performs translation at document-level and can potentially further improve translation quality, e.g., document-level context can help word disambiguation for translating words with multiple senses, document-level translation can help pronoun translation which requires context outside of the current sentence \cite{muller-etal-2018-large}, document-level translation can improve document-level 
lexical cohesion in the translation \cite{voita-etal-2019-good}.

Document-level neural machine translation (NMT) has received much attention in recent years \cite{bao-etal-2021-g,donato-etal-2021-diverse,fernandes-etal-2021-measuring,kang-etal-2020-dynamic,saunders-etal-2020-using,yu-etal-2020-better,ijcai2020p551,yang-etal-2019-enhancing,kuang-etal-2018-modeling,bawden-etal-2018-evaluating, zhang-etal-2018-improving, voita-etal-2018-context, kuang-xiong-2018-fusing}. Existing works showed that document-level translation can outperform sentence-level translation for a number of datasets, such as TED, News,  Europarl \cite{bao-etal-2021-g,donato-etal-2021-diverse,xu-etal-2021-document-graph}. 
Although document-level NMT has shown promising results on a number of benchmarks, document-level NMT is still not widely adopted in real-world translation systems mainly due to the lack of large-scale general domain training data for document-level NMT.

We examine the effectiveness of using Paracrawl \cite{ banon-etal-2020-paracrawl}  for learning document-level NMT. Paracrawl is a large-scale parallel corpus crawled from the Internet and contains data from various domains. The official   Paracrawl corpus\footnote{{https://paracrawl.eu/}} was released as   parallel sentences (extracted from parallel webpages) and therefore previous works   only used Paracrawl   for learning sentence-level translation. 
In this work, we  extract
parallel paragraphs from  Paracrawl parallel webpages using automatic sentence alignments and we use the extracted parallel paragraphs as parallel documents for
 training document-level translation models.   We show that
document-level NMT models trained with only parallel paragraphs from Paracrawl  can be used to  translate real documents from TED, News and Europarl, outperforming sentence-level NMT models and also improving context-aware pronoun
translation \cite{muller-etal-2018-large}.



\section{Extracting Parallel Paragraphs from Paracrawl}
\label{ddata}
Paracrawl \cite{banon-etal-2020-paracrawl} is a large-scale  parallel corpus crawled from the Internet and   contains texts from various domains. The official   Paracrawl corpus was released as   aligned sentence pairs. For high-resource language pairs like German-English, the Paracrawl corpus provides 278M aligned sentences for the translation task. As Paracrawl was released as aligned sentence pairs, previous work only used Paracrawl for learning sentence-level translation. However, as described in the Paracrawl documentation, Paracrawl was constructed by first identifying aligned webpages (URLs) and then aligning sentences within aligned webpages. Although texts in parallel webpages tend to be noisy and
are   only loosely aligned,
we demonstrate that extracting parallel paragraphs  from Paracrawl parallel webpages using automatic sentence alignments can provide effective training data for learning  document-level translation.
Below we describe how we extract parallel paragraphs from Paracrawl.

\paragraph{Extracting} We extract parallel paragraphs from Paracrawl and examine      the effectiveness of using these aligned paragraphs    for learning document-level NMT.
We used German-English, a high-resource language pair, in our experiments.  As the main purpose of the original Paracrawl project is to collect aligned sentences, Paracrawl did not officially release all the webpage-level aligned texts. Paracrawl only released a subset of all the aligned webpage texts,   for the German-English language pair\footnote{{https://www.statmt.org/paracrawl-benchmarks/}}.  We   extract parallel paragraphs from
the released parallel  webpages:
we first download aligned webpages\footnote{https://www.statmt.org/paracrawl-benchmarks/paracrawl-benchmark.en-de.aligned-docs.xz} (one German webpage is aligned with one English webpage) and then download the automatic sentence alignments (vecalign\footnote{https://www.statmt.org/paracrawl-benchmarks/paracrawl-benchmark.en-de.vecalign.xz})  provided by Paracrawl; we then split a webpage into   paragraphs according to the newline symbol; then  extract aligned paragraphs from aligned webpages according to the automatic sentence alignments. 
We consider two paragraphs to be aligned if sentences in these two paragraphs are aligned to each other and not aligned to any other paragraphs.  We discarded   sentences that are not   one-to-one aligned (e.g. one English sentence aligned to two German sentences) and discarded  sentences that are not aligned to any other sentences. We discarded repeated paragraphs and   discarded paragraphs that only contain a single sentence. Paragraphs with non-monotonic sentence alignments were also discarded.

\begin{table}\small
    \centering
    \begin{tabular}{c|c|c|c|c}
    \hline
    &&&\multicolumn{2}{c}{Words}\\
    &Paragraphs  & Sentences  &En&De \\
    \hline
        Train & 1.5M&5.5M &118M&109M\\
         Dev&402&1504&32K&29K\\
         Test&411&1510&33K&30K\\
         \hline
    \end{tabular}
    \caption{Statistics of parallel paragraphs  extracted from Paracrawl. }
    \label{paradata}
\end{table}

 \begin{table}[t]\small
     \centering
     \begin{tabular}{l|l}
     \hline

     \hline
        Original vecalign   & 147M \\
          After parallel paragraph extraction   & 11.7M \\
          After cleaning & 5.5M\\
        \hline
     \end{tabular}
     \caption{Numbers of remaining sentence pairs after different processing steps.}
     \label{remaindata}
 \end{table}

\begin{table}[t]\small
    \centering
    \begin{tabular}{c|c}
    \hline
     Sentences &  Distribution\\

    \hline
        2 & 34.63\% \\
         3&29.31\%\\
         4&15.99\%\\
         
         5$\sim$10&18.29\%\\
         $>$10&1.77\%\\
         \hline
    \end{tabular}
    \caption{Distribution of paragraph length. For example, the first line of numbers mean 34.63\% of the extracted paragraphs contain 2 sentences.}
    \label{slpara}
\end{table}

\paragraph{Cleaning}
To improve the quality of the extracted parallel paragraphs, we    removed sentences which do not belong to the correct language\footnote{https://pypi.org/project/langid/},  removed paragraphs that are too short (contain less than 30 words) and removed paragraphs with more than 50\% overlap. 

 We
randomly split the extracted parallel paragraphs into training, development and test sets as shown in Table~\ref{paradata}. 
Note that, compared to the officially released Paracrawl corpus, the size of parallel paragraphs that we extracted
from Paracrawl is still relatively small. This is because (i) the aligned webpages provided by Paracrawl we used for parallel paragraph extraction is only a subset of all Paracrawl data (ii) many parallel sentences were discarded due to our strict extraction rules as shown in Table~\ref{remaindata}. For future work, we will collect more webpage-level aligned Paracrawl data and test more flexible extraction rules.

 We show the length  statistics  of the parallel paragraphs that we  extracted from Paracrawl in Table~\ref{slpara}. 
  Compared to normal documents, the parallel paragraphs extracted from Paracrawl are generally much shorter. However, sentences in the same paragraph are usually closely related and can  provide useful context information to help the translation of each other. In our experiments, we use   the extracted parallel   paragraphs as parallel documents to train document-level NMT models and we show that  document-level NMT models trained with only parallel paragraphs from Paracrawl  can be used to translate real documents  from TED, News and Europarl, outperforming sentence-level NMT models.

 \section{Document-level Translation with Paracrawl}
\label{expe}
\subsection{Modeling}
Previous works have shown that   document-level NMT models can outperform sentence-level models on several benchmarks.
We adopted the recent document-level translation model, 
G-Transformer  \cite{bao-etal-2021-g}\footnote{{https://github.com/baoguangsheng/g-transformer}}, in our experiments to test the effectiveness of using Paracrawl data for learning document-level translation.
The G-Transformer model is based on the standard sentence-level Transformer \cite{vaswani2017attention}, but uses a whole document together as the input of the model and then generates translation for the whole document. G-Transformer improves the standard Transformer model for document-level translation with extra group tags and group attention. Each word (both source and target words) in the document is assigned with a group tag to indicate which sentence this word belongs to. Compared to the standard Transformer attention, G-Transformer computes group attention using group tags   
       to encourage local attention and  reducing the hypothesis space of the
attention, especially from target to source, for long documents. 
 G-Transformer outperformed sentence-level Transformer and obtained  new
state-of-the-art results
   on three document-level translation
benchmarks.

\subsection{Experimental Setting}

 We tokenize and truecase all data
with MOSES \citep{koehn-etal-2007-moses} scripts, and then perform subword segmentation with
  byte pair encoding   (BPE) \citep{sennrich-etal-2016-neural} using 30k merging
operations.
For both the sentence-level Transformer and the document-level G-Transformer, we used the base model setting \cite{vaswani2017attention} with 6-layer encoder/decoder, 512-dimension word embedding and 2048 hidden units for the feed forward networks.  
Following the G-Transformer experimental settings \cite{bao-etal-2021-g},
 we set the max length of a document to     512 BPE tokens (if a document is longer than 512 tokens, we split it into multiple 
  instances).
  For model training, we first pretrained a standard sentence-level Transformer for 100k training steps and then finetuned the sentence-level Transformer to learn document-level G-Transformer for another 100k training steps. For a fair comparison between sentence-level and document-level translation, we also applied pretraining for sentence-level translation, i.e. we train a sentence-level Transformer with 100k pretraining and 100k finetuning 
 steps in order to compare with G-Transformer.

 \begin{table*}[]\small
     \centering
     \begin{tabular}{ll|ll|ll|ll|ll}
     \hline
    & &\multicolumn{4}{|c|}{Train-Paracrawl-Only}&\multicolumn{4}{|c}{Train-Combined  }\\
     
     &&EnDe&&DeEn&&EnDe&&DeEn\\
      &    &sent &doc&sent&doc&sent&doc&sent&doc \\
      \hline
       BLEU&   Paracrawl& 26.57&\textbf{26.96} &30.94&\textbf{31.94}$^\dagger$ &26.44&\textbf{27.40}$^\dagger$  &31.36&\textbf{31.92}$^\dagger$ \\
        &  Europarl&22.78&\textbf{23.04} &27.76&\textbf{28.69}$^\dagger$ &28.60&\textbf{29.30}$^\dagger$ &35.53&\textbf{35.90}$^\dagger$ \\
         & TED&23.56&\textbf{24.39}$^\dagger$ &27.89&\textbf{28.24} &25.70&\textbf{26.13} &30.15&\textbf{31.49}$^\dagger$ \\
          &News&31.55&\textbf{32.07}$^\dagger$ &34.59&\textbf{35.39}$^\dagger$ &33.08&\textbf{33.89}$^\dagger$ &35.99&\textbf{36.39}\\
          \hline
          COMET&Paracrawl&16.16&\textbf{17.99} &18.90&\textbf{20.99}$^\dagger$ &17.66&\textbf{20.44}$^\dagger$ &20.60&\textbf{22.02}\\
          &Europarl&38.34&\textbf{39.84}$^\dagger$ &42.39&\textbf{45.10}$^\dagger$ &54.10&\textbf{54.80}$^\dagger$ &55.78&\textbf{56.31}$^\dagger$\\
          &TED&23.14&\textbf{23.77} &41.34&\textbf{42.62}&34.21&\textbf{35.25}&46.02&\textbf{46.92}\\
          &News&33.84&\textbf{35.51}$^\dagger$ &44.45&\textbf{48.26}$^\dagger$ &42.49&\textbf{44.20}$^\dagger$ &49.66&\textbf{51.35}$^\dagger$\\
          \hline
     \end{tabular}
     \caption{Translation results.  Train-Paracrawl-Only:  only Paracrawl as training data. Train-Combined: Paracrawl, Europarl, TED and News combined as training data. $^\dagger$ represents a significant difference at the p <
0.01  level.}
     \label{mainres}
 \end{table*}

\begin{table*}[t!]\small
\centering
\begin{tabular}{l|llll|llll}
\hline
&\multicolumn{4}{c}{Train-Paracrawl-only}&\multicolumn{4}{|c}{Train-Combined}\\
 &total &es &er &sie &total &es &er &sie \\
\hline
sent &0.43& 0.91 & 0.15 & 0.24  & 0.43& \textbf{0.94} & 0.17 &0.20 \\ 
doc &\textbf{0.55}& \textbf{0.93}& \textbf{0.34} &\textbf{0.37}  &\textbf{0.60}& 0.92 &\textbf{0.44} &\textbf{0.44}   \\\hline
\end{tabular}

\caption{ Accuracy on contrastive pronoun test set  with regard to reference pronoun. }
\label{parapron}
\end{table*}

\begin{table*}[t!]\small
\centering
\begin{tabular}{l|l|l|l|l}
\hline
&\multicolumn{2}{c|}{Train-Paracrawl-Only}&\multicolumn{2}{c}{Train-Combined}\\
&inside&outside&inside&outside\\
 & current sentence & current sentence &current sentence& current sentence \\
\hline
sent &\textbf{0.68}& 0.37   &0.71 &0.37  \\ 
doc &0.67&\textbf{0.52}  &\textbf{0.76} &\textbf{0.56}   \\\hline
\end{tabular}

\caption{ Accuracy on contrastive pronoun test set with regard to antecedent location. }
\label{paraloca}
\end{table*}

 \subsection{Evaluating with BLEU, COMET  and Targeted Pronoun Evaluation} We used BLEU \cite{papineni2002bleu} and COMET \cite{rei-etal-2020-comet} for translation quality evaluation\footnote{Both BLEU and COMET scores were computed on sentence-level using\\ https://github.com/mjpost/sacrebleu   \\ https://unbabel.github.io/COMET/html/index.html}
and  we performed significance testing   with bootstrap resampling \cite{koehn-2004-statistical}.
In addition to BLEU and COMET which evaluate the general translation quality, we also performed a targeted evaluation for pronoun translation following \citet{muller-etal-2018-large}'s work. The correct translation of a pronoun often requires context outside   of the current sentence. Therefore, evaluation of pronoun translation can demonstrate the advantage of document-level NMT models. Following \citet{muller-etal-2018-large}'s work, we 
computed the accuracy of our models choosing the correct translation for the English pronoun ``it" from the three possible German words ``es", ``er" and ``sie". We used a context of 5 sentences for the test data\footnote{https://github.com/ZurichNLP/ContraPro} provided by \citet{muller-etal-2018-large} for pronoun evaluation.

\begin{table}[]\small
    \centering
    \begin{tabular}{l|l|l|l|l}
    \hline
         & \multicolumn{2}{c|}{T-P-O}&\multicolumn{2}{c}{T-C} \\
         &EnDe&DeEn&EnDe&DeEn\\
         \hline
         Paracrawl&26.88&31.37&27.35&31.65\\
         Europarl&23.22&28.47&29.03&35.61\\
         TED&24.24&28.04&25.62&29.91\\
         News&31.86&34.92&33.41&35.99\\
         \hline
    \end{tabular}
    \caption{Translation results (BLEU) of using the document-level G-Transformer for translating single sentences. T-P-O: Train-Paracrawl-Only. T-C: Train-Combined.}
    \label{bleusingle}
\end{table}

\subsection{Training with Only Paracrawl}
\label{seconlypara}
We trained both sentence-level Transformers and document-level G-Transformers
 with only Paracrawl  (see Table~\ref{paradata}) as training data. Parallel paragraphs were used as parallel documents to train document-level G-Transformers and parallel sentences contained in parallel paragraphs were used to train sentence-level Transformers.\footnote{Therefore,  training data for  the document-level model and the sentence-level model contain the same number of sentence pairs.}
For evaluation, we used test data from
 4 datasets, Paracrawl, Europarl, TED and News. For Europarl, TED and News, we used the same test data following the original G-Transformer  \cite{bao-etal-2021-g} work, i.e., the Europarl, TED and News test sets are parallel documents in contrast to the Paracrawl test set which is parallel paragraphs.
 BLEU and COMET scores are given in Table~\ref{mainres} as Train-Paracrawl-Only.
\begin{table}[t!]\small
    \centering
    \begin{tabular}{l|l|l|l}
    \hline
         \multicolumn{2}{c|}{T-P-O}&\multicolumn{2}{c}{T-C}  \\
         inside&outside&inside&outside \\
         \hline
         0.67&0.36&0.78&0.35\\
         \hline
    \end{tabular}
    \caption{Accuracy on contrastive pronoun test set with regard to antecedent location when using the document-level G-Transformer for translating single sentences without context.}
    \label{prounsingle}
\end{table}
  The document-level G-Transformers  achieved higher translation quality than sentence-level Transformers for all 4 test sets and only used Paracrawl as training data, which demonstrates that Paracrawl can provide useful document-level information  for effective training of document-level NMT models. 
Table~\ref{mainres} also shows that the document-level information contained in Paracrawl is robust across domains as    
 document-level G-Transformers trained with only  Paracrawl data can help to    translate real documents from  TED, News and Europarl test sets.

Table~\ref{parapron} and Table~\ref{paraloca} give results of targeted pronoun evaluation. Results show that the document-level model, trained with only Paracrawl data, significantly outperformed the sentence-level model for pronoun translation especially when the antecedent location of a pronoun is outside of the current sentence (see Table~\ref{paraloca}), which again demonstrates  that Paracrawl can provide useful information outside of the current sentence for effective learning of document-level NMT.

\subsection{Training with Paracrawl, TED, News and Europarl Combined}

We also trained translation models with training data from Paracrawl, Europarl, TED and News combined. The sentence-level Transformers were trained with sentence pairs from all the 4 training datasets. The document-level G-Transformers were trained with parallel paragraphs from Paracrawl and parallel documents from Europarl, TED and News. 
 We then computed BLEU     and COMET scores for the 4 test sets as shown in Table~\ref{mainres} as Train-Combined. The targeted pronoun evaluation results  are also given in Table~\ref{parapron} and Table~\ref{paraloca} as {Train-Combined}.
 Results show that using additional parallel documents together with parallel (Paracrawl) paragraphs as training data   further improved the general translation quality (BLEU and COMET) and also helped the document-level model to obtain a higher accuracy for targeted pronoun translation evaluation.

\subsection{Document-level G-Transformer for Translating Single Sentences}
We also evaluated how the document-level G-Transformer (trained with parallel documents) performs for translating single sentences (i.e. each sentence is considered as a single document  at test time). Table~\ref{bleusingle} shows that, when considering the general translation quality (e.g. BLEU),  document-level G-Transformers can perform well for translating single sentences, even outperforming the sentence-level Transformers in Table~\ref{mainres} for most of the test sets. 
However, the targeted pronoun evaluation results in Table~\ref{prounsingle} show that the accuracy of translating pronouns with antecedent location outside of the current sentence dropped more than 10\% compared to the document-level model in Table~\ref{paraloca}, which demonstrates that document-level G-Transformers indeed  require context outside of the current sentence for improving pronoun translation.


\section{Conclusion      }
 \label{concl}
As document-level translation lacks large-scale general-domain document-level training data, we   examine the effectiveness of using Paracrawl data for learning document-level translation.
 Paracrawl was officially released as parallel sentences extracted from parallel webpages.
In this work, we  extract parallel paragraphs from  Paracrawl aligned webpages   using  automatic sentence alignments 
and we use the extracted parallel paragraphs as parallel documents for training document-level NMT models. We show that document-level  
models trained with only parallel paragraphs from Paracrawl  can be used to translate real documents from
 TED, News and Europarl, outperforming sentence-level Transformers and also improving context-aware pronoun
translation.

 \section*{Limitations}

 Compared to the officially released Paracrawl corpus (278M sentence pairs for German-English), the size of parallel paragraphs  that we extracted from Paracrawl is still relatively small. This is because (i) Paracrawl only released a subset of all webpage-level aligned  texts and we only extracted parallel paragraphs from these released webpage texts (ii) we used very strict rules for extracting parallel paragraphs from Paracrawl and many parallel sentences were discarded by our extraction rules. For future work, we will  collect more Paracrawl webpage-level aligned data for parallel paragraph extraction and we will test more flexible extraction rules.

\section*{Acknowledgements}

The authors acknowledge the financial support by the German Federal Ministry for Education and Research (BMBF) within the projects ``CORA4NLP" {01IW20010} and  ``KI-Servicezentrum Berlin Brandenburg" {01IS22092}.

\bibliography{anthology,custom}
\bibliographystyle{acl_natbib}

\appendix

\end{document}